\documentclass[11pt,a4paper]{article}
\usepackage[hyperref]{acl2020}
\usepackage{times}
\usepackage{latexsym}

\usepackage{url}
\usepackage{booktabs}
\usepackage{multirow}
\usepackage{graphicx}
\usepackage{amsmath}

\usepackage{microtype}

\aclfinalcopy % Uncomment this line for the final submission
\title{Towards Multimodal Response Generation with \\ Exemplar Augmentation and Curriculum Optimization}

\author{Zeyang Lei$^{13*}$, Zekang Li$^{23}$\thanks{Equal Contribution. This work was done when Zeyang Lei and Zekang Li were interning at Pattern Recognition Center, WeChat AI, Tencent.}, Jinchao Zhang$^3$, Fandong Meng$^3$, \\ \textbf{
Yang Feng$^2$, Yujiu Yang$^1$, Cheng Niu$^3$, Jie Zhou$^3$ }\\
  $^1$Tsinghua University, China. \\
  $^2$Key Laboratory of Intelligent Information Processing \\
  Institute of Computing Technology, Chinese Academy of Sciences \\
  $^3$Pattern Recognition Center, WeChat AI, Tencent Inc, China \\
  {\tt \small \{lizekang19,fengyang\}@ict.ac.cn,  yang.yujiu@sz.tsinghua.edu.cn} \\
  {\tt \small zeyanglei@gmail.com, \{dayerzhang,fandongmeng,chengniu,jiezhou\}@tencent.com} \\
  }

\date{}

\begin{document}
\maketitle
\begin{abstract}
 Recently, variational auto-encoder (VAE) based approaches have made impressive progress on improving the diversity of generated responses. However, these methods usually suffer the cost of decreased relevance accompanied by diversity improvements~\cite{zhang2018generating}. In this paper, we propose a novel multimodal\footnote{A multimodal distribution is a continuous probability distribution with two or more modes.} response generation framework with exemplar augmentation and curriculum optimization to enhance relevance and diversity of generated responses. First, unlike existing VAE-based models that usually approximate a simple Gaussian posterior distribution, we present a Gaussian mixture posterior distribution (i.e, multimodal) to further boost response diversity, which helps capture complex semantics of responses. Then, to ensure that relevance does not decrease while diversity increases, we fully exploit similar examples (exemplars) retrieved from the training data into posterior distribution modeling to augment response relevance. Furthermore, to facilitate the convergence of Gaussian mixture prior and posterior distributions, we devise a curriculum optimization strategy to progressively train the model under multiple training criteria from easy to hard. Experimental results on widely used SwitchBoard and DailyDialog datasets demonstrate that our model achieves significant improvements compared to strong baselines in terms of diversity and relevance.
%  \footnote{We will release the source codes of this work after publication for reproducibility.
%  }

%Empowering machines with the ability to converse with human coherently and informatively is a key goal for open-domain dialog systems. Recently,  variational auto-encoder (VAE) based approaches have made impressive progress on this topic. However, existing VAE-based approaches usually approximate the posterior distribution over latent variables using a simple Gaussian distribution, which restricts the ability in capturing complex semantics and high variability of responses. In this paper, we propose a novel dialogue generation architecture based on exemplar augmentation and curriculum optimization to enhance the diversity and relevance of generated responses. Specifically, by fully exploiting similar examples retrieved from the training data, a Gaussian mixture distribution is used to model the posterior distribution of latent variables. Meanwhile, to better approximate Gaussian mixture prior and posterior distributions, a curriculum optimization strategy is devised to progressively train the model under multiple training criteria from easy to hard. Experimental results on widely used SwitchBoard and DailyDialog datasets demonstrate that our model achieves significant improvements compared to strong baselines in terms of diversity and relevance. \footnote{We will release the source code of this work after publication. Our source code is included in Supplementary Material.}
\end{abstract}

\section{Introduction}
Recently, sequence-to-sequence (Seq2Seq) based conversation models have achieved great success in open-domain dialogue generation. However, these methods often generate generic and dull responses~\cite{li2015diversity}, such as \emph{``I don't know", ``It's Ok"}.  A seemingly promising approach is to integrate variational auto-encoders (VAEs)~\cite{serban2017hierarchical,zhao2017learning} or its variants~\cite{du2018variational,gu2018dialogwae} into the encoder-decoder framework to enhance the diversity of generated responses.
%Many researchers attempt to mitigate the issue from different aspects such as modifying the training  objective~\cite{li2015diversity,gao2019jointly}, devising better decoding search schemes~\cite{vijayakumar2016diverse}.
%Recently, open-domain dialogue generation has attracted increasing attention in both industry and academics due to its broad application. It aims to generate coherent and informative responses given the context under different topics. A great deal of approaches \cite{} have been proposed to create open-domain dialogue systems. Among these, Seq2Seq models and its variants \cite{13,Zhou2017Emotional,pandey2018exemplar} have become dominant in literature.
\begin{figure}[t]
		\centering
		\includegraphics[width=0.4\textwidth]{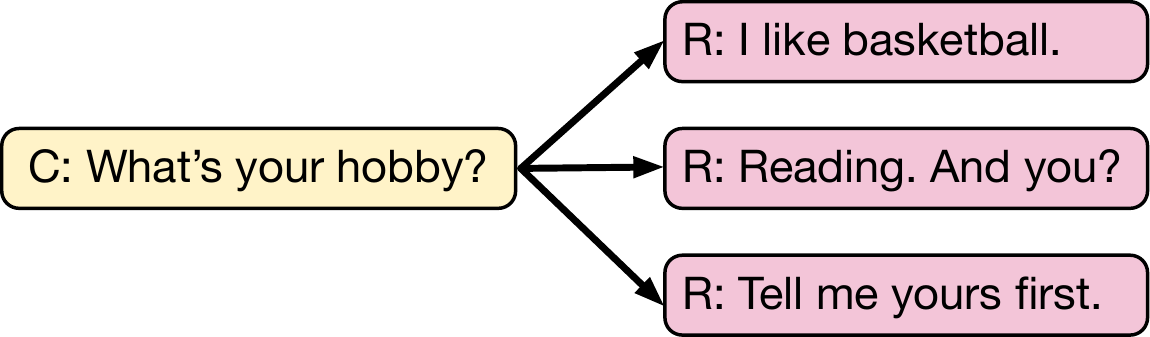}
		\caption{One context might correspond to multiple responses, so Gaussian mixture posterior distribution can help capture high variability of responses. C and R in the figure denote the context and response respectively.} \vspace{-10pt}
		\centering
		\label{fig0}
	\end{figure}

Although existing VAE-based approaches have shown great potential in diverse response generation, these approaches still face two issues. \emph{First}, existing VAE-based approaches usually approximate the posterior distribution over the latent variables using a simple Gaussian distribution, which restricts the ability of these approaches to capture the complex semantics and high variability of responses to some extent. 
\emph{Second}, these approaches usually suffer the cost of decreased relevance accompanied by increased diversity~\cite{zhang2018generating,gao2019jointly}.

To tackle the aforementioned issues, we propose a novel multimodal response generation framework with exemplar augmentation and curriculum optimization to enhance both relevance and diversity of responses. Specifically, to capture the complex semantics of responses, we present a Gaussian mixture posterior distributions to boost the diversity of generating responses. An intuitive explanation about using Gaussian mixture distribution in the posterior distribution is presented in Figure \ref{fig0}. 
Then, to make sure that relevance does not decrease when diversity increases, we fully exploit similar examples (exemplars) retrieved from the training data into Gaussian mixture posterior distribution modeling to augment response relevance. Such motivation is based on that these responses from similar contexts can be regarded as potential exemplar responses for the current context~\cite{pandey2018exemplar,wu2018response}.
Furthermore, to facilitate the convergence of Gaussian mixture prior and posterior distributions, we devise a curriculum optimization strategy to progressively train the model under multiple training criteria from easy to hard. In particular, our model is trained through three phases: firstly training a simple Wasserstein Auto-encoder (WAE)\footnote{As stated in~\cite{zhao2018adversarially}, comparing to KL divergence widely used in conventional VAEs, the Wasserstein distance as a notion of distance may result in a better generative model.} ~\cite{wasserstein2017} only with a simple normal posterior distribution, then training a  complex WAE with multiple simple normal posterior distributions, and finally training our entire model with Gaussian mixture prior and posterior distributions.

The main contributions are as follows: 
\begin{itemize}
\item We propose a Gaussian mixture posterior distribution over the latent variables to capture the high variability of responses. Meanwhile, to ensure that relevance does not decrease when diversity increases, we fully exploit similar examples (exemplars) from the training data in the Gaussian mixture posterior model.
%By exploiting similar examples from the training data, the  posterior  distribution over the latent variables is formulated as a Gaussian mixture model, which helps capture high variability of responses. 
\item  A curriculum optimization strategy is devised to progressively train our model through three phases with training criteria from easy to hard (i.e.,  the convergence of training objectives from easy to hard). %which achieves better diversity and relevance of generated responses. 
%To the best of our knowledge, we are the first to exploit curriculum optimization in open-domain dialogue generation. 
\item Our study shows that: (1)By fully exploiting exemplars, a Gaussian mixture posterior distribution can help improve both diversity and relevance of generated responses; (2) curriculum optimization strategy can facilitate the model training, which further achieves better diversity and relevance of generated responses.

%(1) a Gaussian mixture posterior distribution can help improve response diversity (e.g., intra-dist and inter-dist in the  experiment) ; (2) the use of exemplars can help enhance relevance of generated responses; (3) curriculum optimization strategy can facilitate the model training, which further achieves better diversity and relevance of generated responses.
%\item Experimental results on two popular datasets, SwitchBoard and DailyDialog, demonstrate that our model achieves significant improvements compared to strong baselines in terms of both diversity and relevance.
\end{itemize}

\section{Related Work}
\textbf{Variational Autoencoder (VAE) for Dialogue Generation.}
Recently, some researchers~\cite{bowman2015generating,serban2017hierarchical,zhao2017learning,shen2018improving,park2018hierarchical,fu2019cyclical} have attempted using variational auto-encoders (VAEs) to address the issue that vanilla Seq2Seq models suffer from generating generic and dull responses.  The VAE models introduce latent variables into encoder-decoder frameworks to improve the variability of the models. 
\iffalse
%More formally, suppose that $c=[u_1,u_2,...,u_I]$ refers to a context containing $I$ utterances where $u_i$ denotes an utterance, and $r$ represents a response which is also the next utterance of $u_I$. The goal of general dialogue generation modeling is to learn a conditional distribution $p(r|c)$.
%The variational autoencoder (VAE) estimates the $p(r|c)$ by introducing a latent variable $z$ that denotes a high-level representation of response, shown as follow.
\begin{equation}
p(r|c)=\int_z p(r|c,z)p(z|c)dz
\end{equation}
Specifically, $p(z|c)$ refers to the prior distribution of $z$ given the condition $c$ and can be implemented with a neural network called as \emph{prior network}. Meanwhile, the posterior distribution $q(z|r,c)$ of $z$ is modeled with another neural network named \emph{recognition network} to approximate the true posterior distribution $p(z|r,c)$. The VAE model is trained to maximize the likelihood of a response by computing the evidence lower bound (ELBO)~\cite{sohn2015learning}:
\begin{align}
   L(r,c)&=-{\rm{KL}}(q(z|r,c)|p(z|c))+ \nonumber \\ 
   &\mathbf{E}_{z\sim q(z|r,c)}[{\rm{log}} p(r|c,z)] \leq {\rm{log}}p(r|c). 
\end{align}
\fi
Most existing VAEs based models in dialogue generation usually used a simple Gaussian model for the prior and posterior distribution. This restricts the ability in capturing complex semantics and high variability of context and responses.  \citet{gu2018dialogwae} proposed a Gaussian mixture \emph{prior} to enrich the latent space.  Compared with \cite{gu2018dialogwae}, our model propose Gaussian mixture \emph{posterior} distributions over the latent variables to capture complex semantics of responses. Meanwhile, we utilizes similar examples retrieved from training data in posterior distribution modeling, which can better approximate the true posterior distribution and generate more related responses. 
\iffalse
Significantly, compared with \cite{gu2018dialogwae} as a latest work, our model mainly differs in three aspects: (i) While \citet{gu2018dialogwae} modeled a Gaussian mixture \textbf{prior} distribution, its \textbf{posterior} distribution is still a simple normal distribution. Our model proposes a Gaussian mixture \textbf{posterior} distribution to widen the variability of posterior distribution; (ii) For the implementation of Gaussian mixture distribution, \citet{gu2018dialogwae} adopted a Gumbel-softmax re-parameterization to compute a multimodal distribution with energy almost \textbf{concentrated on one single peak}. However, our model models Gaussian mixture \textbf{prior} distribution by using softmax function and Gaussian mixture \textbf{posterior} distribution by using the weighted sum of multiple normal distributions, which can approximate more complex semantics distribution (e.g. multimodal with multiple arbitrary peaks). More details will be shown in section \textbf{Methodology}; (iii) As the matching between Gaussian mixture prior and posterior distributions involves more parameters and thus is difficult to train, we propose a curriculum optimization strategy to facilitate the training of model.
\fi

\begin{figure*}[t]
		\centering
		\includegraphics[width=0.85 \textwidth]{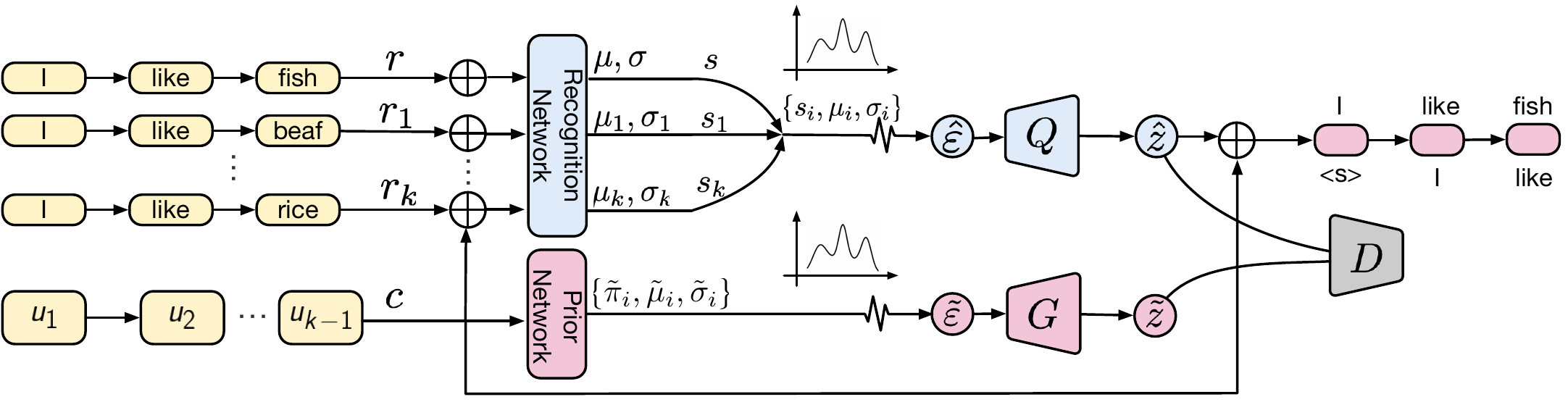}
		\caption{The architecture of our proposed model. $\oplus$ denotes the concatenation of the input vectors. Q and G represent two generators, and D is a discriminator. Q, G and D are used to measure the Wasserstein distance between prior and posterior distribution.} \vspace{-10pt}
		\centering
		\label{model overview}
	\end{figure*}

\noindent \textbf{Curriculum learning.}
Curriculum learning is a machine learning strategy, which starts from simple subtasks and then gradually handles harder ones~\cite{bengio2009curriculum}. The learning strategy has been proven effective in many NLP tasks. For instance, \citet{liu2018curriculum} utilized curriculum learning to solve the natural answer generation problem by firstly learning models on low-quality question-answer (QA) pairs and then on high-quality QA pairs. Meanwhile, some researchers~\cite{zhang2018empirical,Emmanoui2019curriculum} used curriculum learning to enhance the neural machine translation (NMT) by choosing the samples from easy to hard according to certain criteria. Inspired by such ideas, we propose a curriculum optimization strategy to better train our model under multiple training criteria from easy to hard. Unlike conventional curriculum learning that uses samples from easy to hard, our proposed curriculum optimization strategy is set based on the convergence of training objectives from easy to hard (e.g., in this paper, firstly training a simple WAE model to learn the basic encoder and decoder, then training a complex WAE model to fully learn \emph{recognition network}, and finally training the entire model with Gaussian mixture prior and posterior distributions until convergence). 

\section{Methodology}
% \subsection{Model Overview}
Figure \ref{model overview} demonstrates an overview of our model. Our model mainly contains Prior and Recognition Network, Wasserstein GAN (i.e., Q, G, D), and basic encoder and decoder.

\subsection{Overview}  
At training stage, we first input the current context and each response pair (including the golden and retrieved similar responses) to the utterance and context encoder to obtain the corresponding hidden representations, and then feed them to a shared feed-forward network named \emph{recognition network} to obtain the mean and covariance of a normal distribution for each response. 
Each response-context pair corresponds to a simple Gaussian distribution and then we compute the  Gaussian mixture distribution weighted by the similarity between the real context and the retrieved similar context. Next, we use a re-parameterization trick to draw a Gaussian mixture noise from
the \emph{recognition network}. Finally, we employ a generator Q to transform the posterior Gaussian noise into a sample of the posterior latent variable. 

Similarly, the output of the \emph{prior network} is also a Gaussian mixture distribution to match the prior distribution with posterior distribution better. In particular, we use a feed-forward network as \emph{prior network} to transform the context into the means and covariances of the corresponding Gaussian components. Then a prior Gaussian noise is sampled from the \emph{prior network} and fed to a generator G to obtain a sample of the prior latent variable. 

Finally, we introduce an adversarial discriminator D to match the posterior distribution with the prior distribution by minimizing the Wasserstein distance between them. At the generation stage, the decoder RNN takes as inputs the prior latent variable and the context to generate a response. In the following, we will elaborate our model via two sections including \textbf{Exemplar-augmented Conditional Wasserstein Auto-encoders} and \textbf{Curriculum Optimization}.

\subsection{Exemplar-augmented Conditional Wasserstein Auto-encoders}
Given a context-response pair ($c$, $r$), the similar examples $(c_{i},r_{i}), i=1,2,...,k$ can be obtained by using the last utterance of the context $c$ as a query to retrieve from the training data using the BM25~\cite{robertson2009probabilistic} retrieval model. 
Then we use the utterance encoder and the context encoder which both adopt gated recurrent units (GRUs) to encode context or responses into fix-sized vectors. In particular, the utterance encoder encodes each utterance into a fixed-sized vector, and the context encoder takes as input the encoding vector of the preceding utterance and uses the final hidden state $h(c)$ of the context encoder as the context representation. Thus, we can obtain the context and response representation  $(h(c_i),h(r_i))$ 
% (For convenience of description, we use ($c_0$, $r_0$) to represent ($c$, $r$) in this paper).

\noindent \textbf{(i) Prior and Recognition Network.}
Different from previous VAE-based models, both the prior and posterior distribution of our model are Gaussian mixture distributions. The posterior distribution is a Gaussian mixture distribution explicitly composed of multiple simple Gaussian distributions conditioned on the exemplar responses and the gold response. 
Specificly, the posterior latent variable $\hat{z}$$\sim$$Q_\phi(\hat{\epsilon})$ is generated by a  generator Q from a context-response-dependent Gaussian mixture noise $\hat{\epsilon}$, which is a reparametrization trick~\cite{kingma2013auto}. 
And $\hat{\epsilon}$ is sampled from a Gaussian mixture distribution which is composed of $k$ Gaussian simple distribution whose mean $\mu_i$ and covariance $\sigma_i^2$ can be calculated by a feed-forward neural network named \emph{recognition network} (RecNet) as follows:
\begin{equation}
    \setlength{\abovedisplayskip}{4pt} 
	\setlength{\belowdisplayskip}{4pt}
\begin{split}
\label{eq:prigen}
\small
\hat{z}=Q_\phi(\hat{\epsilon}), ~~~ \hat{\epsilon} \sim \sum_{i=0}^{k} s_{i}\mathcal{N}(\epsilon_i; \mu_i, \sigma^2_iI) \\ \begin{bmatrix} \mu_i \\ \log\sigma^2_i      \end{bmatrix} = W f_\phi(\begin{bmatrix} h(c) \\ h(r_i) \end{bmatrix}) + b
% s_i=\frac{exp(cos(h(c), h(c_i)))}{\sum_{i=0}^k exp(cos(h(c), h(c_i)))} \\
% cos(h(c),h(c_i)) = \frac{h(c) \cdot h(c_{i})}{\Vert h(c)\Vert \Vert h(c_{i})\Vert}
\end{split}
\end{equation}
where $f_\phi$ represents a feed-forward neural network and $W$, $b$ refer to the trainable parameters. For convenience of description, we use ($c_0$, $r_0$) to represent ($c$, $r$) in this paper. The weight score of each Gaussian simple distribution can be computed as follows:

\begin{figure*}[t]
		\centering
		\includegraphics[width=0.8 \textwidth]{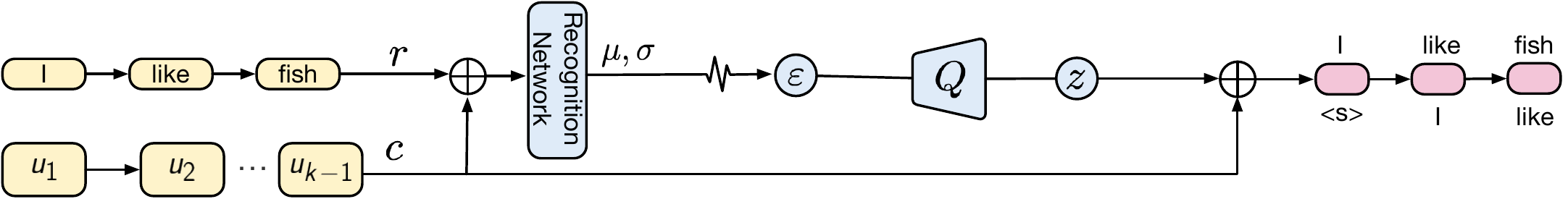}
		\caption{Step I in Curriculum Optimization with a simple WAE}
		\centering
		\label{curriculum-1}
	\end{figure*}
\begin{figure*}[t]
		\centering
		\includegraphics[width=0.8 \textwidth]{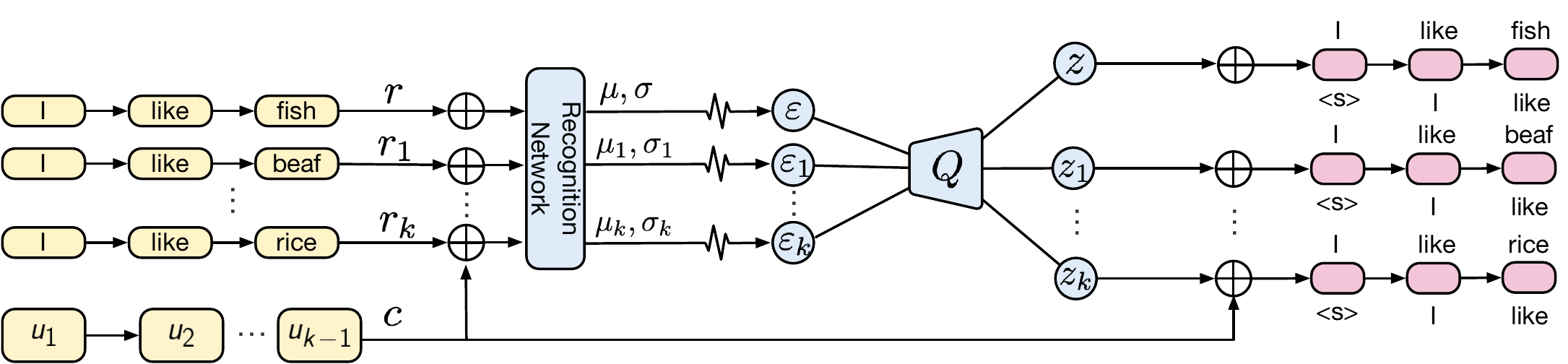}
		\caption{Step II in Curriculum Optimization with a complex WAE} \vspace{-10pt}
		\centering
		\label{curriculum-2}
	\end{figure*}
	
\begin{equation}
    \setlength{\abovedisplayskip}{4pt} 
	\setlength{\belowdisplayskip}{4pt}
\begin{split}
s_i=\frac{exp(cos(h(c), h(c_i)))}{\sum_{i=0}^k exp(cos(h(c), h(c_i)))} \\
cos(h(c),h(c_i)) = \frac{h(c) \cdot h(c_{i})}{\Vert h(c)\Vert \Vert h(c_{i})\Vert}
 \label{s}
\end{split}
\end{equation}
where $cos(h(c),h(c_i))$ denotes the cosine similarity between $h(c)$ and $h(c_i)$ , and $s_i$ represents the normalized weight score of the Gaussian simple distribution $\mathcal{N}(\epsilon_i; \mu_i, \sigma^2_iI)$.

Similarly, the prior sample $\tilde{z}$$\sim$$p(\tilde{z}|c)$ can be generated by a generator G from a context-dependent random noise $\tilde{\epsilon}$. $\tilde{\epsilon}$ is also drawn from a Gaussian mixture distribution composed of $n$ simple Gaussian components over the context, which can be computed by a feed-forward neural network named \emph{prior network} (PriNet) as follows:
\begin{equation}
\begin{split}
\small
\tilde{z}=G_\theta(&\tilde{\epsilon}), ~~~ \tilde{\epsilon}\sim\sum_{i=1}^n\tilde{\pi}_i\mathcal{N}(\tilde{\epsilon}; \tilde{\mu}_i, \tilde{\sigma}^2_iI) \\ &\tilde{\pi}_i=\frac{exp(\alpha_i)}{\sum_{i=1}^n exp(\alpha_i)} \\
&\begin{bmatrix} \alpha_i \\ \tilde{\mu}_i \\ \log\tilde{\sigma}^2_i      \end{bmatrix} = \tilde{W}_i g_\theta(h(c)) + \tilde{b}_i\\
\end{split}
\end{equation}
where $g_\theta$ represents a feed-forward neural network. $\tilde{W}_i$ and $\tilde{b}_i$ denotes the learnable parameters. 

%From the view of the implementation, unlike \citep{gu2018dialogwae}, our prior distribution use $softmax$ instead of $Gumble-softmax$ to compute the weights of each Gaussian distribution component, which benefits approximating more complex Gaussian mixture distribution.

\noindent \textbf{(ii) Wasserstein GAN.}
Meanwhile, to alleviate the posterior collapse problem~\cite{shen2018improving}, we match the Gaussian mixture prior and approximate posterior distribution by using WGAN~\cite{arjovsky2017wasserstein} to minimize the Wasserstein distance between them, which have been shown to produce good results in text generation ~\cite{zhao2018adversarially,gu2018dialogwae}.
Formally, our model can be trained by maximizing:
\begin{align}
    L(c,R)&=-W(q(\hat{z}|c,R)|p(\tilde{z}|c))+ \nonumber \\ 
   &\mathbf{E}_{\hat{z}\sim q(\hat{z}|c,R)}[\log p(r|c,\hat{z})]
\end{align}
where $W(\cdot|\cdot)$ denotes the Wasserstein distance between the two distributions, $p(r|c,\hat{z})$ represents a decoder RNN and $R=[r,r_1,...,r_k]$. The detailed theory and implementation about Wasserstein distance can be found in \citep{arjovsky2017wasserstein}.

\subsection{Curriculum Optimization}
% \begin{figure*}[t]
% 		\centering
% 		\includegraphics[width=1.0\textwidth]{Figures/curriculum-3.pdf}
% 		\caption{Step 3 in Curriculum Learning}
% 		\centering
% 		\label{curriculum-3}
% 	\end{figure*}
To further facilitate the matching between Gaussian mixture prior distribution and posterior distributions to better train our model, we devise a curriculum optimization strategy containing three phases: training a Wasserstein Auto-encoder (WAE) with a single normal posterior distribution by using the gold response, training a WAE with multiple normal posterior distributions by using multiple exemplar responses, and training our entire model with Gaussian mixture prior and posterior distributions by using all the exemplar responses and the gold response. The training procedure gradually increases difficulty with training criteria from easy to hard. It is noted that the WAE in the first and second phases does not contain prior distribution compared with conventional WAE.

Specifically, in the first phase presented in Figure \ref{curriculum-1}, we train our model by minimizing the reconstruction loss only over the gold response as follows:
\begin{equation}
\mathcal{L}_{1} = -\mathbf{E}_{z=Q_\phi(\epsilon),\;\epsilon\sim \mathrm{RecNet}(c,r)}\log p_\psi(r|c, z)
\end{equation}
The training objective of this phase is to obtain a better encoder-and-decoder, warming up for the following phase. In the second  phase presented in Figure \ref{curriculum-2}, we train our model on the basis of the first phase by minimizing the reconstruction loss over multiple exemplar responses as follows:
\begin{align} 
\small
 \mathcal{L}_{2} &= -  \nonumber \mathbf{E}_{z_i=Q_\phi(\epsilon_i),\;\epsilon_i\sim\mathrm{RecNet}(c,r_i)} \xi ,\\
 & \xi =\sum_{i=1}^k s(h(c),h(c_i))\log p_\psi(r_i|c, z_i)
\end{align}
where $s(h(c),h(c_i))$ is the above weight score in Equation \ref{s}. This phase can ensure that each component of the complex Gaussian mixture distribution to be fully trained for the total posterior distribution.
In the final WAE phase, we train our entire model shown in Figure \ref{model overview} by minimizing the discriminator loss from the discriminator D computed as follows:
\begin{align}
\mathcal{L}_{disc} = \mathbf{E}_{\hat{\epsilon}\sim \mathrm{RecNet}( c,r)}[D(Q(\hat{\epsilon}),c)] - \nonumber \\ 
 \mathbf{E}_{\tilde{\epsilon}\sim \mathrm{PriNet}(c)}[D(G(\tilde{\epsilon}),c)]
\end{align}
The final total loss function for the third phase can be computed as follow:
\begin{align}
    \mathcal{L}_3=-\mathbf{E}_{\hat{z}\sim q(\hat{z}|c,R)}[{\rm{log}} p(x|c,\hat{z})]+\mathcal{L}_{disc}
\end{align}
Through these three phases, we can further improve the diversity and relevance of generated responses.

\section{Experiment}

\begin{table*}[tb]
		\centering
		\begin{tabular}{l@{}|c@{~~}c@{~~}c| c@{~~}c@{~~}c | c@{~~}c | c@{~~}c}
			\hline
			\multirow{2}{*}{\bf Model} & \multicolumn{3}{c|}{\bf BLEU} & \multicolumn{3}{c|}{\bf BOW Embedding} & \multicolumn{2}{c|}{\bf intra-dist} & \multicolumn{2}{c}{\bf inter-dist}\\ \cline{2-11}
			& R & P & F1 &  G &  E  &  A  &  dist-1 &  dist-2 & dist-1 & dist-2 \\ \hline			
			HRED & 0.232 & 0.232 & 0.232 & 0.798  ~& 0.511  ~&  0.915 ~&  \bf 0.935  ~&  0.969  ~& 0.093  ~&  0.097\\
            SeqGAN & 0.270 & \textbf{0.270} & 0.270 &0.774  ~& 0.495  ~&  0.907 ~& 0.747  ~&  0.806 ~&  0.075 ~&   0.081 \\
			%CVAE & 0.265 & 0.222 & 0.242 & 0.811  ~& 0.543  ~&  0.923 ~&  0.938  ~&  0.973 ~&0.177  ~&  0.222\\
			%CVAE-BOW & 0.256 & 0.224 & 0.251 & 0.812  ~& 0.540  ~&  0.923 ~&\bf  0.947 ~&  \bf 0.976  ~&0.165 ~&  0.206\\
			CVAE-CO & 0.259 & 0.244 & 0.251 & 0.818  ~& 0.530  ~& 0.914 ~&  0.821  ~&  0.911 ~&0.106 ~&  0.126 \\
            VHRED  & 0.271 & 0.260 & 0.265 & 0.786  ~& 0.507  ~&  0.892 ~& 0.633  ~&  0.771 ~&0.071 ~&  0.089 \\
            VHCR & 0.289 & 0.266 & 0.277 & 0.798  ~& 0.525  ~&  0.925 ~& 0.768  ~&  0.814 ~&0.105 ~&  0.129 \\
			DialogWAE-GMP & 0.336 & 0.230 & 0.273 & \bf 0.861 ~&  0.597 ~&  0.945~&  0.881 ~&  0.946~& 0.450 ~&  0.776 \\ % epoch 80
			\hline
			Our model & \bf 0.357 & 0.255 & \bf 0.297 &  0.851 ~&  \bf 0.599 ~&  \bf 0.951 ~&  0.883 ~& \bf 0.978~& \bf 0.479 ~& \bf 0.843\\  % epoch 70
			\hline
			~~ w/o I & 0.325 & 0.208 & 0.254 & 0.856  ~& 0.589  ~&  0.942 ~& 0.922  ~&  0.980 ~& 0.594 ~&  0.911 \\ %epoch 70 
			~~ w/o II & 0.343 & 0.234 & 0.278 & 0.863  ~& 0.610  ~&  0.948 ~& 0.903  ~&  0.972 ~& 0.507 ~&  0.844  \\ %epoch 70
			~~ w/o Curriculum & 0.304 & 0.180 & 0.226 & 0.867  ~& 0.611  ~&  0.941 ~& 0.850  ~&  0.890 ~& 0.516 ~&  0.801  \\ %epoch 70
			~~ w/o examplar & 0.289 & 0.188 & 0.228 & 0.850 & 0.617 & 0.940 & 0.951 & 0.983 & 0.449 & 0.686  \\
			\hline
		\end{tabular}
		\caption{Performance comparison on the DailyDialog dataset (G: Greedy, E: Extrema, A: Average)}
		\label{tab:eval:acc:dd}
	\end{table*}

\begin{table*}[tb]
		\centering
		\begin{tabular}{l@{}|c@{~~}c@{~~}c| c@{~~}c@{~~}c | c@{~~}c | c@{~~}c}
			\hline
			\multirow{2}{*}{\bf Model} & \multicolumn{3}{c|}{\bf BLEU} & \multicolumn{3}{c|}{\bf BOW Embedding} & \multicolumn{2}{c|}{\bf intra-dist} & \multicolumn{2}{c}{\bf inter-dist} \\ \cline{2-11}
			& R & P & F1 & G &  E  &  A &  dist-1 &  dist-2 & dist-1 & dist-2 \\ \hline
			
			HRED & 0.262 & 0.262 & 0.262 & 0.832  ~& 0.537  ~&  0.820  ~& 0.813  ~&  0.452 ~& 0.081  ~&  0.045   \\
            SeqGAN & 0.282 & 0.282 & 0.282  & 0.748 ~& 0.515  ~&  0.817 ~& 0.705  ~&  0.521 ~& 0.070 ~&  0.052  \\
			%CVAE & 0.295 & 0.258 & 0.275 & 0.846  ~& 0.572  ~& 0.836 ~& 0.803  ~&  0.415 ~&0.112  ~&  0.102\\
			%CVAE-BOW & 0.298 & 0.272 & 0.284 & 0.840  ~& 0.555  ~&  0.828 ~& 0.819  ~&  0.493 ~&0.107  ~&  0.099\\
			CVAE-CO & 0.299 & 0.269 & 0.283 &0.855  ~& 0.557  ~&  0.839 ~& 0.863  ~&  0.581 ~&0.111 ~&  0.110  \\
            VHRED & 0.253 & 0.231 & 0.242  &0.844  ~& 0.531  ~&  0.810 ~& \bf 0.881  ~&  0.522 ~&0.110 ~&  0.092  \\
            VHCR & 0.276 & 0.234 & 0.254 & 0.851  ~& 0.546  ~&  0.826 ~& 0.877  ~&  0.536 ~&0.130 ~&  0.131  \\
			DialogWAE-GMP  & \bf 0.411 & \bf 0.241 & \bf 0.304 & \bf 0.893 ~& \bf 0.657 ~&  \bf 0.918~& 0.805  ~& 0.704 ~& 0.384  ~&  0.648 \\ %epoch 40
			\hline
			Our model & \bf 0.410 & \bf 0.240 & \bf 0.303 & \bf 0.893 ~&   0.650 ~& \bf 0.918 ~& 0.823 ~& \bf 0.780 ~&\bf 0.440 ~&\bf 0.707  \\
			\hline
			~~ w/o I & 0.383 & 0.221 & 0.280 & 0.888  ~& 0.636  ~&  0.909 ~& 0.849  ~&  0.763 ~&0.528 ~&  0.801  \\ %epoch 
			~~ w/o II & 0.403 & 0.217 & 0.282 & 0.895  ~& 0.668  ~&  0.919 ~& 0.810  ~&  0.637 ~&0.394 ~&  0.615 \\ %epoch 70 
			~~ w/o Curriculum & 0.402 & 0.217 & 0.282 & 0.894  ~& 0.664  ~&  0.916 ~& 0.799  ~&  0.611 ~&0.385 ~&  0.605  \\ %epoch 60 
			~~ w/o examplar & 0.425 & 0.236 & 0.304 & 0.892 & 0.670 & 0.922 & 0.753 & 0.583 & 0.269 & 0.389   \\ %epoch 70
			\hline
		\end{tabular}
		\caption{Performance comparison on the SwitchBoard dataset (G: Greedy, E: Extrema, A: Average)} \vspace{-10pt}
		\label{tab:eval:acc:sw}
		%\ {-1\baselineskip}
	\end{table*}
	
\subsection{Datasets}
We conduct experimWents on two widely used dialogue datasets~\cite{zhao2017learning,shen2018improving}, SwitchBoard~\cite{godfrey1997switchboard} and DailyDialog~\cite{li2017dailydialog}. We split the datasets into training, validation and test sets by the same ratios as the baselines methods, that is, 2316:60:62 for Switchboard~\cite{zhao2017learning} and 10:1:1 for Dialydialog~\cite{shen2018improving}, respectively.

%DailyDialog holds 13118 daily conversations crawled from different websites for English learners. Each conversation includes about 8 speaker turns, and 15 tokens per utterance. Switchboard contains 2,400 two-side telephone conversations from 70 specified topics. 

\subsection{Baselines}
We carefully select the following six related state-of-the-art methods as baselines:

\noindent  \noindent \textbf{HRED}:  a generative hierarchical encoder-decoder network \citep{serban2016building}. 

\noindent \textbf{SeqGAN}: a GAN model for dialogue generation \cite{li2017adversarial}.

%\noindent \textbf{CVAE}: a conditional VAE model with KL-annealing for dialogue modeling \citep{zhao2017learning}.

%\noindent \textbf{CVAE-BOW}: a conditional VAE model a Bow loss~\citep{zhao2017learning}.

\noindent \textbf{CVAE-CO}: a collaborative conditional VAE model \citep{shen2018improving}.

\noindent \textbf{VHRED}: a hierarchical encoder-decoder framework with VAE \cite{serban2017hierarchical}.

\noindent \textbf{VHCR}: a hierarchical VAE model with conversation modeling \cite{park2018hierarchical}.

\noindent \textbf{DialogWAE-GMP}: a conditional Wasserstein autoencoder (WAE) with Gaussian mixture prior network for dialogue modeling (DialogWAE-GMP) \cite{gu2018dialogwae}. We rerun its released source codes with default parameters.

\subsection{Metrics}
\begin{figure*}[t]
		\centering
		\includegraphics[width=1.0\textwidth]{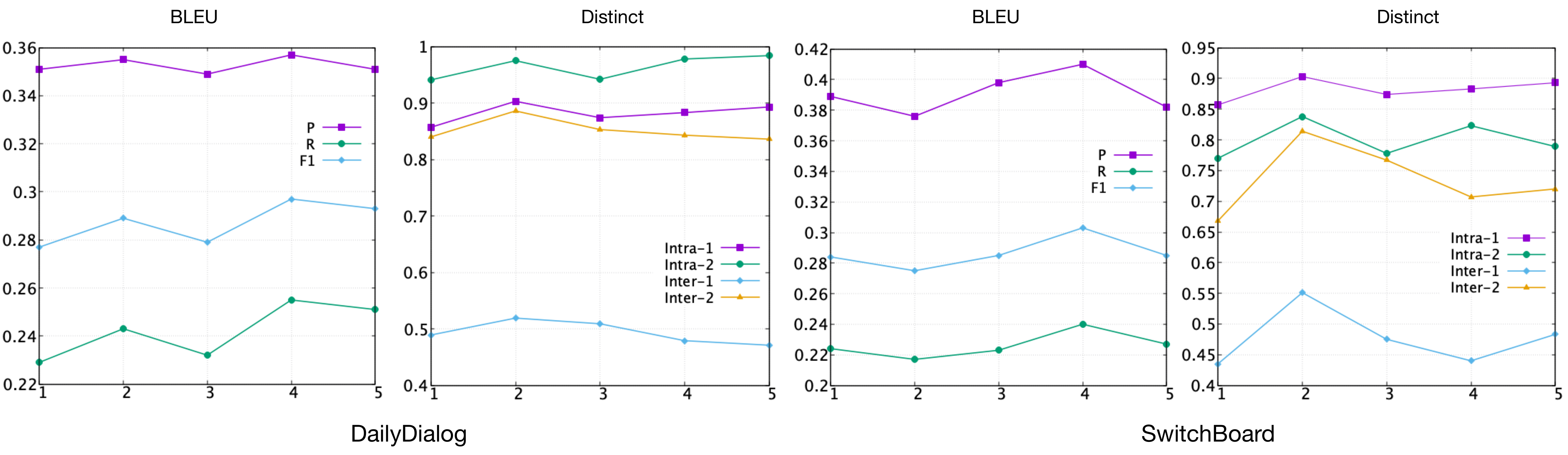}
		\caption{Performance with respect to the number of examplar cases}
		\centering
		\label{DailyDial-k}
	\end{figure*}

\begin{table}[t]
        \centering
        \small
        \begin{tabular}{l|c|c|c } 
        \hline
        Model & Fluency & Relevance & Diversity \\ 
        \hline
        DialogWAE & 24.1\% & 27.3\% & 22.5\% \\ 
        Our model & {\bf 44.0\%} & {\bf 38.2\%} & {\bf 46.3\%} \\ 
        w/o curriculum &  31.9\% & 34.5\% & 31.2\% \\
        \hline
        \end{tabular}
        \caption{Human evaluation on the DailyDialog dataset}
        \label{tab:eval:human}
    \end{table}
 
 \begin{table}[t]
        \centering
        \small
        \begin{tabular}{l|c|c|c } 
        \hline
        Model & Fluency & Relevance & Diversity \\ 
        \hline
        DialogWAE & 20.3\% & 23.3\% & 23.4\% \\ 
        Our model & {\bf 48.2\%} & {\bf 46.2\%} & {\bf 44.3\%} \\ 
        w/o curriculum &  31.5\% & 30.5\% & 32.3\% \\
        \hline
        \end{tabular}
        \caption{Human evaluation on the SwitchBoard dataset.} \vspace{-10pt}
        \label{tab:eval:human_swichboard}
    \end{table}

    \begin{table}[t]
        \centering
        \small
        \begin{tabular}{l|c|c|c } 
        \hline
        Model & Fluency & Relevance & Diversity \\ 
        \hline
        Our model & {\bf50.6} \% & {\bf 48.7}\% & {\bf 48.7}\% \\ 
        ~~~w/o I &  20.7\% &  7.3\% &  22.0\% \\ 
        ~~~w/o II &  18.0\% & 21.3\% & 18.0\% \\
        ~~~w/o curriculum &  10.7\% & 22.7\% & 11.3\% \\
        \hline
        \end{tabular}
        \caption{Ablation study on the DailyDialog dataset}
        \label{tab:eval:ablation}
    \end{table}

    \begin{table}[t]
        \centering
        \small
        \begin{tabular}{l|c|c|c } 
        \hline
        Model & Fluency & Relevance & Diversity \\ 
        \hline
        Our model & {\bf 56.0\%} & {\bf 48.0\%} & {\bf 51.0\%} \\ 
        ~~~w/o I &  20.1\% &  14.7\% &  23.8\% \\ 
        ~~~w/o II &  15.9\% & 19.2\% & 15.9\% \\
        ~~~w/o curriculum &  8.0\% & 18.1\% & 10.3\% \\
        \hline
        \end{tabular}
        \caption{Ablation study on the SwitchBoard dataset.}
        \label{tab:eval:ablation_switchboard}
    \end{table}

\begin{table*}[t]
\small
		\centering
		\begin{tabular}{p{3cm}|lp{4.5cm}|lp{4.5cm}} 
		    \hline
			\textbf{Context} & \multicolumn{2}{c|}{\textbf{DialogWAE-GMP}} & \multicolumn{2}{c}{\textbf{Our model}} \\ \cline{1-5}
			
			\multirow{3}{3cm}{how much is the rent? $\rightarrow$ the rent is \$1500 per month.} & \textbf{Eg}.1: & no problem. i'll take it. & \textbf{Eg}.1: & oh, that's too expensive.  \\ 
            
            & \textbf{Eg}.2: & this one is \$1.50.50,000 yuan per month. & \textbf{Eg}.2: & will it be cash dollars?\\
            
			&\textbf{Eg}.3: & that sounds like a good idea. & \textbf{Eg}.3: & i have to pay for it. \\
			
			\hline
			
			\multirow{3}{3cm}{guess who i saw just now? $\rightarrow$ who? $\rightarrow$ john smith. $\rightarrow$ that bad egg who took the low road since he was a boy.} & \textbf{Eg}.1: & it is my favorite. & \textbf{Eg}.1: & his face is lost.  \\ 
            
            & \textbf{Eg}.2: & no, but i didn't think he was able to get married. i had no idea to get her. & \textbf{Eg}.2: & oh, he's really very serious. he doesn't look like the one.\\
            
			&\textbf{Eg}.3: & this is not, but it’s not that bad. it’s just a little bit, but it’s not too bad. & \textbf{Eg}.3: & it must be right around the moment. \\
			\hline
			
            \multirow{3}{3cm}{hello. is bob there, please? $\rightarrow$ hold on a second please. i'm sorry. he has gone out. $\rightarrow$ oh. this is paul. when will he be back? } & \textbf{Eg}.1: & no. not at all. i'm sorry, but i'm not sure. & \textbf{Eg}.1: & please come to the office tomorrow afternoon. he will be there for 6:30.\\ 
            
            & \textbf{Eg}.2: & yes, but he has a number of the company. he is a little bit. & \textbf{Eg}.2: & of course. he said he would be able to call his secretary to call him for next week.\\
            
			&\textbf{Eg}.3: & no, i don't think so. it's the first time to get a new one. & \textbf{Eg}.3:  & please, sir, the doctor will be back in the morning. \\
		    
		    \hline
		\end{tabular}
		\caption{Examples of context-response pairs for the neural network models. $\rightarrow$ indicates a change of turn. `Eg.i' means the $i$-th example.} \vspace{-10pt}
		\label{tab:eval:eg}
	\end{table*}
%We evaluate the proposed model from two aspects: automatic evaluation and human evaluation. 

\noindent \textbf{Automatic Evaluation.} 
%In  dialogue generation, it is  widely debated that how well BLEU score against a single reference can reflect the quality of generated response~\cite{liu2016not,serban2017hierarchical}. Thus, we do not use BLEU scores in our experiments as in \cite{serban2017hierarchical,pandey2018exemplar}. Instead, 
To evaluate our model, we adopt three widely used metrics that can reflect the relevance and diversity: BLEU, BOW Embedding and distinct. BLEU measures how much a generated response contains n-gram overlaps with the reference.
We use smoothing techniques to compute BLEU scores for $n<4$ \cite{chen2014systematic}.
BOW Embedding represents the cosine similarity of bag-of-words embeddings between the predicted and gold responses, which has been used in many studies~\cite{du2018variational,gu2018dialogwae} to evaluate the relevance of generated responses. In this paper, we use three commonly used BOW Embedding metrics including Greedy~\cite{rus2012comparison}, Extrema~\cite{forgues2014bootstrapping} and Average~\cite{mitchell2008vector}. In the test stage, we sample 10 predicted responses for each test context and compute the maximum BOW embedding score among 10 sampled responses as the final reported results.
\iffalse
\textbf{Greedy} represents greedily computing the maximum of the cosine similarities between the word embeddings in two sentences, and to average the obtained values. \textbf{Average} refers to that the cosine similarity is computed based on the averaged word embeddings in the two sentences. \textbf{Extrema} denotes computing the cosine similarity between the largest extreme values among the word embeddings of the two sentences. 
\fi
Distinct measures the diversity of generated responses. \emph{dist-n} computes the fraction of distinct n-grams (n=1,2) among all n-grams in generated responses~\cite{li2015diversity}. We compute the \emph{intra-dist} as the average of distinct values within each sampled response and \emph{inter-dist} as the distinct value among all the sampled responses.
%BLEU 

\noindent \textbf{Human Evaluation.}
As human evaluation is essential for dialogue generation, we randomly sampled 150 dialogues from the test set of DailyDialog and Switchboard to conduct a human evaluation. For each context in the test, we generated 10 responses from evaluated models. Responses for each context were inspected by 3 annotators who were asked to choose the model which performs the best in regards \emph{Fluency}, \emph{Relevance} and \emph{Diversity} among all the compared models. Finally, the ratio of each model under each metric was computed as the corresponding human evaluation score.
\emph{Fluency} means that how likely generated responses are produced by a human. \emph{Relevance} means that how likely generated response is relevant to the context. \emph{Diversity} means that how much generated responses provide specific information rather than dull and meaningless information. To ensure the fairness of evaluation, the evaluation was conducted in a strictly random and blind fashion to rule out human bias. 

\subsection{Experiment Settings}
In our experiment, the utterance encoder is a bidirectional GRU~\cite{pennington2014glove}, and both the context encoder and decoder are vanalia GRUs. The hidden size of all GRUs is set to 300. The prior and the recognition networks are both 2-layer feed-forward networks of size 200 with tanh non-linearity. The generators Q and G as well as the discriminator D are 3-layer feed-forward networks with ReLU non-linearity. The dimension of all latent variables is set to 200. We adopt pre-trained Glove~\cite{pennington2014glove} with a size of 200 as the word embedding. We use the RMSprop optimizer with a mini-batch size of 32. The epochs for the first two curriculum optimization phases are 10 and 10 respectively. In this paper, to simplify the settings, we adopt the same number of prior and posterior components.
\subsection{Experiment Results}
\noindent \textbf{Automatic Evaluation Results.} As shown in Table \ref{tab:eval:acc:dd} and \ref{tab:eval:acc:sw}, our model outperforms all the baselines in most of automatic metrics on the two datasets, especially in inter/intra-dist. To be specific, for DialyDialog, our model achieves the highest BLEU R/F1 scores and BOW Extrema/Average scores compared to all baselines, indicating that our model can enhance \textbf{relevance} of generated responses. Meanwhile, our model obtains significant improvements in terms of inter-dist and intra-dist on DialyDialog, indicating that our model can enhance response \textbf{diversity}. For SwitchBoard, our model achieves the highest inter-dist and intra-dist (dist-2) scores , and comparable BLEU and BOW Embedding scores with the state-of-the-art method (DialogWAE-GMP). This confirms that our model can also improve relevance when diversity significantly increases. 

%for BOW embedding metrics, our model obtains the highest Greedy and Extrema scores among all baselines with achieving very approximate Average score compared with the state-of-the-art DialogWAE-GMP.
%This means that our model can generate more coherent responses compared to strong baselines.
%For \emph{intra-dist}, our model also achieves comparable performance with the start-of-the-art methods, which 
%show that our model can obtain comparable intra-sentence diversity and generate more informative sentences. 
%In terms of \emph{inter-dist}, our model significantly outperforms all the baselines by a large margin. For example, our model gains 78\% and 48\% improvements in dist-1 and dist-2 compared to DialogWAE-GMP for SwitchBoard.
%This indicates that our model can enhance inter-sentence diversity and contains more distinct $n$-grams.

\noindent \textbf{Human Evaluation Results.} To further evaluate our model, we conduct a human evaluation on the state-of-the-art method--- DialogWAE-GMP (denoted as DialogWAE), our model and our model without curriculum optimization (denoted as w/o curriculum). The results\footnote{The Kappa values for inter-annotator agreement are 0.8 on two datasets, showing good agreement.} are shown in Table \ref{tab:eval:human} and \ref{tab:eval:human_swichboard}. Our model significantly and consistently outperforms DialogWAE-GMP in terms of fluency, relevance, and diversity, indicating that our model can enhance both \textbf{relevance} and \textbf{diversity}. Concretely, our model w/o curriculum optimization (that is, with exemplars augmentation and Gaussian mixture posterior) obtains better diversity and relevance than DialogWAE-GMP (with a simple Gaussian posterior distribution). This indicates that \textbf{Gaussian mixture posterior distribution modeled on exemplars} can help improve the diversity and relevance of generated responses.
Meanwhile, our model significantly outperforms our model w/o curriculum in terms of all metrics, which indicates \textbf{curriculum optimization} can further improve the diversity and relevance of generated responses. 
\subsection{Quantitative Analysis}
\noindent \textbf{Ablation Study.} To analyze the effectiveness of each component of our model, we conduct an ablation study on two datasets by automatic and human evaluation. Specifically, we discard the first phase of curriculum optimization (denoted as w/o I), the second phase (denoted as w/o II), the entire curriculum optimization (denoted as w/o curriculum), and exemplars\footnote{Here, removing exemplars means that modeling a Gaussian mixture posterior distribution on a single gold response.}(denoted as w/o exemplar). The automatic evaluation results are shown in Table \ref{tab:eval:acc:dd} and \ref{tab:eval:acc:sw}. The human evaluation results are shown in Table \ref{tab:eval:ablation} and Table \ref{tab:eval:ablation_switchboard}. The The Kappa values is 0.61 and 0.67 on two datasets respectively.

According to automatic and human evaluation results, the relevance or diversity goes down when our model removes each part, indicating that each phase of curriculum optimization and exemplars make contributions to the diversity and relevance of our model. Specifically, for our model w/o I (or II), the relevance decreases, indicating that I and
II can help improve the relevance of responses. For our model w/o curriculum, both its diversity and relevance go down, indicating that \textbf{curriculum optimization} can benefit both diversity and relevance. For our model w/o exemplars, its relevance goes down on DailyDialog compared with DialogWAE, indicating that \textbf{exemplars} can help improve the relevance of responses. 
Meanwhile, for our model w/o exemplars, its diversity increases on Dailydialog compared with DialogWAE, indicating that \textbf{Gaussian mixture posterior distribution} can help improve the diversity of generated responses to some extent. An interesting finding is that though automatic metrics results (BLEU and BOW) on SwitchBoard seems to be good, its generated responses contain large amounts of low-quality responses with many repetitive words or phrases, which will results in poor human evaluation.

%Thus this confirms that modeling Gaussian mixture posterior distribution in exemplars is an effective method to boost both relevance and diverse of generated responses.
%Meanwhile, since our model w/o exemplar has Gaussian mixture posterior distribution (based on a single response) and  DialogWAE has simple posterior distribution, this verifies Gaussian mixture posterior distribution can help improve diversity of generated responses. 

\noindent \textbf{The Exemplar Number (k).} To further investigate the effect of the exemplar number (i.e., the number of posterior components), we conduct the experiments on two datasets by varying k from 1 to 5. The performance under different k is shown in Figure \ref{DailyDial-k}. From the figure, we can conclude that, in most cases, the performance increases with k and decreases once k exceeds a certain threshold (e.g., k=4). 
% Moreover, we also found that, with k varying, the BLEU have relatively small fluctuation while the \emph{distinct} values are sensitive to k.  This indicates that the relevance is basically stable with k increasing and we can improve the diversity at no loss of relevance by choosing the proper k value. 
Considering comprehensively the relevance and diversity of generated responses, we set k to 4 in our experiment. 

\subsection{Case Study} 
To empirically analyze the quality of generated responses, 
we present some examples generated from our model and the start-of-the-art DialogWAE-GMP in Table \ref{tab:eval:eg}. For each context, we show three samples of generated responses from each model. It can be seen that our model generates more relevant, fluent and diverse responses than DialogWAE-GMP. Specifically, for the first context, all the samples generated from our model are related to the topic \emph{``how much"} while the responses Eg.1 and Eg.3 from DialogWAE-GMP seems not very relevant with the topic, which indicates that the responses generated by our model have better \textbf{relevance} than DialogWAE-GMP. Meanwhile, we can also observe that responses from DialogWAE-GMP have a certain token repetition phenomenon. For instance, in the second case, the token ``it's not that bad" emerges two times in the response Eg.3 of DialogWAE-GMP  while such phenomenon have not been found in our model, which shows that our model can generate more \textbf{fluent} and human-like responses. Finally, from the third case, we can see that the responses generated by our model contain more specific information than the responses from DialogWAE (including some meaningless safe responses, e.g., ``I am not sure''). This confirms that our model can generate more \textbf{diverse} and informative responses.

%As human evaluation is relatively accurate and reasonable for evaluating the quality of the generated responses~\citep{}, we carry out a human evaluation to validate the effectiveness of our model and each component of our model.

\section{Conclusion}
In this paper, we propose a novel multimodal response generation framework with exemplar augmentation and curriculum optimization to enhance the diversity and relevance of generated responses. In specific, we first fully exploit exemplars to approximate more complex Gaussian mixture distribution, which is helpful for modeling the high variability of generated responses. Meanwhile, we progressively train our model with curriculum optimization through three phases with training criteria from easy to hard, which facilitates model training to further improve the diversity and relevance of responses.
The experimental results on two popular datasets demonstrate our model can generate more diverse and relevant responses compared with strong competitors.

\bibliography{anthology,acl2020}
\bibliographystyle{acl_natbib}

%\clearpage
\appendix{}
%\section{Variational Autoencoder (VAE)}
%Suppose that $c=[u_1,u_2,...,u_I]$ refers to a context containing $I$ utterances where $u_i$ denotes an utterance, and $r$ represents a response which is also the next utterance of $u_I$. The goal of general dialogue generation modeling is to learn a conditional distribution $p(r|c)$.
%The variational autoencoder (VAE) estimates the $p(r|c)$ by introducing a latent variable $z$ that denotes a high-level representation of response, shown as follow.
%\begin{equation}
%(r|c)=\int_z p(r|c,z)p(z|c)dz
%\end{equation}
%Specifically, $p(z|c)$ refers to the prior distribution of $z$ given the condition $c$ and can be implemented with a neural network called as \emph{prior network}. Meanwhile, the posterior distribution $q(z|r,c)$ of $z$ is modeled with another neural network named \emph{recognition network} to approximate the true posterior distribution $p(z|r,c)$. The VAE model is trained to maximize the likelihood of a response by computing the evidence lower bound (ELBO)~\cite{sohn2015learning}:
%\begin{align}
 %  L(r,c)&=-{\rm{KL}}(q(z|r,c)|p(z|c))+ \nonumber \\ 
 %  &\mathbf{E}_{z\sim q(z|r,c)}[{\rm{log}} p(r|c,z)] \leq {\rm{log}}p(r|c). 
%\end{align}

\end{document}